# Region-Based Incremental Pruning for POMDPs

**Zhengzhu Feng**
Computer Science Department
University of Massachusetts
Amherst, MA 01003
fengzz@cs.umass.edu

**Shlomo Zilberstein**
Computer Science Department
University of Massachusetts
Amherst, MA 01003
shlomo@cs.umass.edu

## Abstract

We present a major improvement to the incremental pruning algorithm for solving partially observable Markov decision processes. Our technique targets the cross-sum step of the dynamic programming (DP) update, a key source of complexity in POMDP algorithms. Instead of reasoning about the whole belief space when pruning the cross-sums, our algorithm divides the belief space into smaller regions and performs independent pruning in each region. We evaluate the benefits of the new technique both analytically and experimentally, and show that it produces very significant performance gains. The results contribute to the scalability of POMDP algorithms to domains that cannot be handled by the best existing techniques.

## 1   INTRODUCTION

A partially observable Markov decision process (POMDP) models an agent acting in an uncertain environment, equipped with imperfect actuators and noisy sensors. It provides an elegant and expressive framework for modeling a wide range of problems in decision making under uncertainty. However, this expressiveness in modeling comes with a prohibitive computational cost when it comes to solving a POMDP and obtaining an optimal policy. Improving the scalability of solution methods for POMDPs is thus a critical research topic.

Standard solution methods for POMDPs rely on performing a dynamic programming update of the value function, represented by a finite set of linear vectors over the state space. A key source of complexity is the size of the value function representation, which grows exponentially with the number of observations. Fortunately, a large number of vectors in this representation can be pruned away without affecting the values. There is a standard linear programming (LP) method for detecting these useless vectors. Solving the resulting large number of linear programs is therefore the main computation in the DP update.

Consequently, many research efforts have focused on improving the efficiency of pruning useless vectors. Among the state-of-the-art techniques, incremental pruning (IP) (Zhang & Liu 1996; Cassandra, Littman, & Zhang 1997; Cassandra 1998) has proved to be particularly efficient. Since its introduction, it has been used as a basic component of several new POMDP algorithms including an improved policy iteration (PI) algorithm (Hansen 1998), symbolic dynamic programming (Hansen & Feng 2000), and a symbolic approximation algorithm (Feng & Hansen 2001), just to name a few. These algorithms use IP as a component and exploit various structures in a typical POMDP for performance gains. For many of these algorithms, the DP update, and hence IP, remains a dominant factor that affects their performance.

We introduce a major improvement to the basic incremental pruning technique, originally proposed by Cassandra *et. al* (Cassandra, Littman, & Zhang 1997; Cassandra 1998). In particular, our work can be seen as a generalization of the *restricted region* (RR) variant of incremental pruning. The RR algorithm exploits the special structure in the cross-sum of *two sets* of vectors to reduce the number of constraints in the LPs. Because of the two-set restriction, the effectiveness of the RR algorithm diminishes as more sets are cross-summed in the incremental pruning process. We show how to overcome this restriction so that the kind of structure exploited by RR can be extended to the whole cross-sum process. The resulting algorithm preserves the simplicity of the original IP technique. Yet, it delivers superb performance improvements. It also preserves the generality of the original IP algorithm, and can thus be embedded into the more advanced algorithms cited above.

The rest of the paper is organized as follows. In the next section, we briefly review the POMDP model and the standard dynamic programming algorithm for solving it. Section 3 reviews IP. Section 4 reviews the RR variant, but from a slightly different perspective than the original work to facilitate the discussion of our algorithms. Section 5 presents and analyzes our main algorithms. We present experimental results in section 6, and conclude the paper in section 7.

## 2 PARTIALLY OBSERVABLE MARKOV DECISION PROCESSES

We consider a discrete time POMDP defined by the tuple $(S, A, P, R, Z, O)$, where

- $S$ is a finite set of states;
- $A$ is a finite set of actions.
- $P$ is the transition model, $P^a(s'|s)$ is the probability of reaching state $s'$ if action $a$ is taken in state $s$;
- $R$ is the reward model, $R^a(s)$ is the expected immediate reward for taking action $a$ in state $s$;
- $Z$ is a finite set of observations that the agent can actually sense;
- $O$ is the observation model, $O^a(z|s')$ is the probability that observation $z$ is seen if action $a$ is taken and resulted in state $s'$.

We are interested in maximizing the infinite horizon total discounted reward, where $\beta \in [0, 1)$ is the discount factor. The standard approach to solving a POMDP is to convert it to a *belief-state* MDP. A belief state $b$ is a probability distribution over the state space $b : S \to [0, 1]$, such that $\sum_{s \in S} b(s) = 1.0$. Given a belief state $b$, representing the agent's current estimate of the underlying states, the next belief state $b'$ is the revised estimate as a result of taking action $a$ and receiving observation $z$. It can be computed using Bayesian conditioning as follows:

$$b'(s') = \frac{1}{P^a(z|b)} O^a(z|s') \sum_{s \in S} P^a(s'|s) b(s),$$

where $P^a(z|b)$ is a normalizing factor:

$$P^a(z|b) = \sum_{s' \in S} \left[ O^a(z|s') \sum_{s \in S} P^a(s'|s) b(s) \right]$$

We use $b' = T(b)$ to refer to belief update. It has be shown that a belief state updated this way is a sufficient statistic that summarizes the entire history of the process. It is the only information needed to perform optimally. An equivalent, completely observable MDP, can be defined over this belief state space as the tuple $(\mathcal{B}, A, T, R_\mathcal{B})$, where $\mathcal{B}$ is the infinite space of all belief states, $A$ is the action set as before, $T$ is the belief transition function as defined above, and $R_\mathcal{B}$ is the reward model, constructed from the POMDP model: $R_\mathcal{B}^a(b) = \sum_{s \in S} b(s) R^a(s)$.

In this form, a POMDP can be solved by iteration of a *dynamic programming update* that improves a value function $V : \mathcal{B} \to \Re$. For all belief states $b \in \mathcal{B}$:

$$V'(b) = \max_{a \in A} \left\{ R_\mathcal{B}^a(b) + \beta \sum_{z \in Z} P^a(z|b) V(T(b)) \right\}. \quad (1)$$

Performing the DP update is challenging because the space of belief states is continuous. However, Smallwood and Sondik (Smallwood & Sondik 1973) proved that the DP backup preserves the piecewise linearity and convexity of the value function, leading the way to designing POMDP algorithms. A piecewise linear and convex value function $V$ can be represented by a finite set of $|S|$-dimensional vectors of real numbers, $\mathcal{V} = \{v^0, v^1, \ldots, v^k\}$, such that the value of each belief state $b$ is defined by

$$V(b) = \max_{v^i \in \mathcal{V}} b \cdot v^i,$$

where $b \cdot v := \sum_{s \in S} b(s) v(s)$ is the "dot product" between a belief state and a vector. Moreover, a piecewise linear and convex value function has a unique minimal-size set of vectors that represents it. This representation of the value function allows the DP update to be computed exactly. Among several algorithms that have been developed to perform this DP step, incremental pruning (IP) is considered the most efficient.

## 3 INCREMENTAL PRUNING

In their description of incremental pruning, Cassandra *et al.* (Cassandra, Littman, & Zhang 1997) note that the updated value function $V'$ of Equation (1) can be defined as a combination of simpler value functions:

$$\begin{aligned} V'(b) &= \max_{a \in A} V^a(b) \\ V^a(b) &= \sum_{z \in Z} V^{a,z}(b) \\ V^{a,z}(b) &= \frac{R_\mathcal{B}^a(b)}{|Z|} + \beta P^a(z|b) V(T(b)) \end{aligned}$$

Each of these value functions is piecewise linear and convex, and can be represented by a unique minimum-size set of vectors. We use the symbols $\mathcal{V}'$, $\mathcal{V}^a$, and $\mathcal{V}^{a,z}$ to refer to these minimum-size sets.

Using the script letters $\mathcal{U}$ and $\mathcal{W}$ to denote sets of vectors, we adopt the following notation to refer to operations on sets of vectors. The *cross sum* of two sets of vectors, $\mathcal{U}$ and $\mathcal{W}$, is defined by $\mathcal{U} \oplus \mathcal{W} = \{u + w | u \in \mathcal{U}, w \in \mathcal{W}\}$. An operator that takes a set of vectors $\mathcal{U}$ and reduces it to its unique minimum form is denoted $\mathbb{PR}(\mathcal{U})$. We also use $\mathbb{PR}(\mathcal{U})$ to denote the resulting minimum set. Formally, $u \in \mathbb{PR}(\mathcal{U})$ if and only if $u \in \mathcal{U}$, and $\exists b \in \mathcal{B}$ such that for $\forall u' \neq u \in \mathcal{U}, u \cdot b > u' \cdot b$.

Table 1: Algorithm for pruning a set of vectors $\mathcal{W}$.

```
procedure POINTWISE-DOMINATE(w, U)
1.  for each u ∈ U
2.      if w(s) ≤ u(s), ∀s ∈ S then return true
3.  return false
procedure LP-DOMINATE(w, U)
4.  solve the following linear program
        variables: d, b(s) ∀s ∈ S
        maximize d
        subject to the constraints
            b · (w − u) ≥ d, ∀u ∈ U
            ∑_{s∈S} b(s) = 1
5.  if d ≥ 0 then return b
6.  else return nil
procedure BEST(b, U)
7.  max ← −∞
8.  for each u ∈ U
9.      if (b · u > max) or ((b · u = max) and (u <_{lex} w))
10.         w ← u
11.         max ← b · u
12. return w
procedure ℙℝ(W)
13. D ← ∅
14. while W ≠ ∅
15.     w ← any element in W
16.     if POINTWISE-DOMINATE(w, D) = true
17.         W ← W − {w}
18.     else
19.         b ← LP-DOMINATE(w, D)
20.         if b = nil then
21.             W ← W − {w}
22.         else
23.             w ← BEST(b, W)
24.             D ← D ∪ {w}
25.             W ← W − {w}
26. return D
```

Using this notation, the minimum-size sets of vectors defined earlier can be computed as follows:

$$\mathcal{V}' = \mathbb{PR}\left(\cup_{a \in A} \mathcal{V}^a\right) \tag{2}$$
$$\mathcal{V}^a = \mathbb{PR}\left(\oplus_{z \in Z} \mathcal{V}^{a,z}\right) \tag{3}$$
$$\mathcal{V}^{a,z} = \mathbb{PR}\left(\{v^{a,z,i} | v^i \in \mathcal{V}\}\right), \tag{4}$$

where $v^{a,z,i}$ is the vector computed by

$$v^{a,z,i}(s) = \frac{R^a(s)}{|Z|} + \beta \sum_{s' \in S} O^a(z|s') P^a(s'|s) v^i(s').$$

Table 1 summarizes an algorithm, due to White and Lark (White 1991), that reduces a set of vectors to a unique, minimal-size set by removing "dominated" vectors, that is, vectors that can be removed without affecting the value of any belief state.

There are two tests for dominated vectors. The simplest method of removing dominated vectors is to remove any vector $u$ that is pointwise dominated by another vector $w$. That is, $u(s) \leq w(s)$ for all $s \in S$. The procedure POINTWISE-DOMINATE in Table 1 performs this operation. Although this method of detecting dominated vectors is fast, it cannot detect all dominated vectors.

There is a linear programming method that can detect all dominated vectors. Given a vector $w$ and a set of vectors $\mathcal{U}$ that does not include $v$, the linear program in procedure LP-DOMINATE of Table 1 determines whether adding $w$ to $\mathcal{U}$ improves the value function represented by $\mathcal{U}$ for any belief state $b$. If it does, the variable $d$ optimized by the linear program is the maximum amount by which the value function is improved, and $b$ is the belief state that optimizes $d$. If it does not, that is, if $d \leq 0$, then $w$ is dominated by $\mathcal{U}$.

The algorithm summarized in Table 1 uses these two tests for dominated vectors to prune a set of vectors to its minimum size. The symbol $<_{lex}$ in the pseudo-code denotes lexicographic ordering. Its significance in implementing this algorithm was elucidated by Littman (1994).

Among the three pruning steps, Equations (2) and (4) can be carried out relatively efficiently with respect to their input size. Equation (3) presents a major bottle-neck because the size of the cross-sum is the product of the inputs: $|\mathcal{U} \oplus \mathcal{W}| = |\mathcal{U}| \times |\mathcal{W}|$. As a result, it is necessary to process $\prod_z |\mathcal{V}^{a,z}|$ vectors in computing $\mathcal{V}^a$. This translates into solving $\prod_z |\mathcal{V}^{a,z}|$ LPs. Incremental pruning addresses this problem by exploiting the fact that the $\mathbb{PR}$ and $\oplus$ operators can be interleaved:

$$\mathbb{PR}(\mathcal{U} \oplus \mathcal{V} \oplus \mathcal{W}) = \mathbb{PR}(\mathcal{U} \oplus \mathbb{PR}(\mathcal{V} \oplus \mathcal{W})). \tag{5}$$

Thus Equation (3) can be computed as follows:

$$\mathcal{V}^a = \mathbb{PR}(\mathcal{V}^{a,z_1} \oplus \mathbb{PR}(\mathcal{V}^{a,z_2} \oplus \cdots \mathbb{PR}(\mathcal{V}^{a,z_{k-1}} \oplus \mathcal{V}^{a,z_k}) \cdots)), \tag{6}$$

which is what the IP algorithm does. The benefit of IP is the reduction of the number of LPs needed to be solved. This can best be understood when Equation (6) is viewed as a recursive process: Instead of pruning the cross-sum $\oplus_{z \in Z} \mathcal{V}^{a,z}$ directly, IP breaks it down by recursively computing $\mathbb{PR}(\oplus_{i=2}^k \mathcal{V}^{a,z_i})$ first, and then prune $\mathcal{V}^{a,z_1} \oplus \mathbb{PR}(\oplus_{i=2}^k \mathcal{V}^{a,z_i})$. Because the size of $\mathbb{PR}(\oplus_{i=2}^k \mathcal{V}^{a,z_i})$ can potentially be much smaller than $\prod_{i=2}^k |\mathcal{V}^{a,z_i}|$, the number of LPs needed to prune $\oplus_{z \in Z} \mathcal{V}^{a,z}$ is reduced from $\prod_z |\mathcal{V}^{a,z}|$ to $|\mathcal{V}^{a,z_1}| \times |\mathbb{PR}(\oplus_{i=2}^k \mathcal{V}^{a,z_k})|$. Note that this argument applies equally to the recursive step $\mathbb{PR}(\oplus_{i=2}^k \mathcal{V}^{a,z_i})$. In general, the total number of LPs used by IP and its variants in computing Equation (3) is asymptotically $|\mathcal{V}^a| \sum_z |\mathcal{V}^{a,z}|$.

## 4 PRUNING OF $(\mathcal{U} \oplus \mathcal{W})$

To prune a single cross-sum of two vector sets $\mathcal{U}$ and $\mathcal{W}$, incremental pruning first constructs the whole set $\mathcal{U} \oplus \mathcal{W}$, and then uses the $\mathbb{PR}$ procedure (Table 1) to prune it. This requires $|\mathcal{U}| \times |\mathcal{W}|$ LPs; each LP has a number of constraints

Table 2: Pruning of $\mathcal{U} \oplus \mathcal{W}$ via region intersection.

**procedure** LP-INTERSECT($\mathcal{B}_\mathcal{U}^u, \mathcal{B}_\mathcal{W}^w$)
1. construct the following linear program:
   variables: $b(s)\ \forall s \in S$
   maximize 0
   subject to the constraints
   $\quad b \cdot (u - u') > 0,\ \forall u' \in \mathcal{U} - \{u\}$
   $\quad b \cdot (w - w') > 0,\ \forall w' \in \mathcal{W} - \{w\}$
   $\quad \sum_{s \in S} b(s) = 1$
2. if the linear program is feasible, return **TRUE**
3. else return **FALSE**

**procedure** $\mathbb{PR}^*(\mathcal{U} \oplus \mathcal{W})$
4. $\mathcal{K} \leftarrow \phi$
5. for each $(u + w) \in \mathcal{U} \oplus \mathcal{W}$
6. $\quad$ if LP-INTERSECT($\mathcal{B}_\mathcal{U}^u, \mathcal{B}_\mathcal{W}^w$)
7. $\quad\quad \mathcal{K} \leftarrow \mathcal{K} \cup \{(u + w)\}$
8. return $\mathcal{K}$

Table 3: Restricted region pruning.

**procedure** LP-DOMINATE($\mathcal{B}_\mathcal{U}^u, w, \mathcal{W}$)
1. solve the following linear program
   variables: $d,\ b(s)\ \forall s \in S$
   maximize d
   subject to the constraints
   $\quad b \cdot (u - u') > 0,\ \forall u' \in \mathcal{U} - \{u\}$
   $\quad b \cdot (w - w') > d,\ \forall w' \in \mathcal{W}$
   $\quad \sum_{s \in S} b(s) = 1$
2. if $d > 0$ return $b$
3. else return nil

**procedure** $\mathbb{PR}(\mathcal{B}_\mathcal{U}^u, \mathcal{W})$
4. $\mathcal{D} \leftarrow \emptyset$
5. while $\mathcal{W} \neq \emptyset$
6. $\quad w \leftarrow$ any element in $\mathcal{W}$
7. $\quad b \leftarrow$ LP-DOMINATE($\mathcal{B}_\mathcal{U}^u, w, \mathcal{D}$)
8. $\quad$ if $b = nil$ then
9. $\quad\quad \mathcal{W} \leftarrow \mathcal{W} - \{w\}$
10. $\quad$ else
11. $\quad\quad w \leftarrow$ BEST($b, \mathcal{W}$)
12. $\quad\quad \mathcal{D} \leftarrow \mathcal{D} \cup \{w\}$
13. $\quad\quad \mathcal{W} \leftarrow \mathcal{W} - \{w\}$
14. return $\mathcal{D}$

ranging from 1 to $|\mathbb{PR}(\mathcal{U} \oplus \mathcal{W})|$, which in the worst case can be $|\mathcal{U}| \times |\mathcal{W}|$[1]. Cassandra *et. al* (Cassandra, Littman, & Zhang 1997; Cassandra 1998) noticed that the set $\mathcal{U} \oplus \mathcal{W}$ has a special structure, and developed the *restricted region* (RR) algorithm to take advantage of this structures to reduce the number of constraints in the LPs. In this section we review their RR algorithm, but from two different angles that will facilitate the discussion of our new algorithms. We note that these two views build on the original intuitions behind the RR algorithm; they have been described less formally in the original paper.

### 4.1 PRUNING BY REGION INTERSECTION

Assume both $\mathcal{U}$ and $\mathcal{W}$ are minimal, in other words, $\mathcal{U} = \mathbb{PR}(\mathcal{U})$ and $\mathcal{W} = \mathbb{PR}(\mathcal{W})$. In addition, assume that all vector sets involved in a cross-sum contain at least 2 vectors, since the cross-sum with a single vector does not introduce any new dominated vector. Each vector $u \in \mathcal{U}$ defines a *witness region* $\mathcal{B}_\mathcal{U}^u$ over which $u$ dominates all other vectors in $\mathcal{U}$ (Littman, Cassandra, & Kaelbling 1996):

$$\mathcal{B}_\mathcal{U}^u = \{b | b \cdot (u - u') > 0, \forall u' \in \mathcal{U} - \{u\}\}. \quad (7)$$

For simplicity of notation, we use $\mathcal{B}_\mathcal{U}$ to refer to a belief region defined by some vector in $\mathcal{U}$, when the specific vector is irrelevant or understood from the context. We also use $\tilde{\mathcal{B}}$ to refer to some region when the vector and vector set are irrelevant or understood from the context.

Note that each inequality in Equation (7) can be represented by a vector, $(u - u')$, over the state space. We call the inequality associated with such a vector a *region constraint*, and use the notation $\mathbb{L}(\mathcal{B}_\mathcal{U}^u) := \{(u - u') | u' \in \mathcal{U} - \{u\}\}$ to represent the set of region constraints defining $\mathcal{B}_\mathcal{U}^u$. Note that for any two regions $\mathcal{B}_\mathcal{U}^u$ and $\mathcal{B}_\mathcal{W}^w$,

$$\mathbb{L}(\mathcal{B}_\mathcal{U}^u \cap \mathcal{B}_\mathcal{W}^w) = \mathbb{L}(\mathcal{B}_\mathcal{U}^u) \cup \mathbb{L}(\mathcal{B}_\mathcal{W}^w). \quad (8)$$

---
[1] The exact number of constraints or LPs here and throughout the paper may vary slightly from the asymptotic figures we use.

The RR algorithm exploits a special structure in the cross-sum set $\mathcal{U} \oplus \mathcal{W}$, captured in the following theorem. (For space reason, all theorems are stated without proof). Let $\phi$ stand for the empty set.

**Theorem 1** *Let $u \in \mathcal{U}$ and $w \in \mathcal{W}$. Then $(u + w) \in \mathbb{PR}(\mathcal{U} \oplus \mathcal{W})$ if and only if $\mathcal{B}_\mathcal{U}^u \cap \mathcal{B}_\mathcal{W}^w \neq \phi$.*

Therefore, computing $\mathbb{PR}(\mathcal{U} \oplus \mathcal{W})$ is equivalent to finding all $u$ and $w$ such that $\mathcal{B}_U^u \cap \mathcal{B}_W^w \neq \phi$. Given $\mathcal{B}_\mathcal{U}^u$ and $\mathcal{B}_\mathcal{W}^w$, the linear program in the procedure LP-INTERSECT listed in Table 2 can be used to determine if they intersect. We call this LP the *intersection* LP. The procedure $\mathbb{PR}^*$ then uses this linear program to construct the minimum set $\mathbb{PR}(\mathcal{U} \oplus \mathcal{W})$. We call $\mathbb{PR}^*$ the *intersection-based* pruning operator.

Compared to using the original $\mathbb{PR}$ operator (Table 1) to directly prune $\mathcal{U} \oplus \mathcal{W}$, the intersection-based pruning requires the same number of LPs to be solved, but each LP has a fixed number of constraints, $|\mathcal{U}| + |\mathcal{W}|$. On one hand, the $\mathbb{PR}$ procedure requires at worst $|\mathbb{PR}(\mathcal{U} \oplus \mathcal{W})|$ constraints. On the other hand, there are also some LPs in $\mathbb{PR}$ that require less than $|\mathcal{U}| + |\mathcal{W}|$ constraints. Because of this, Cassandra *et. al* (Cassandra, Littman, & Zhang 1997; Cassandra 1998) developed a clever way of exploiting the above structure without relying on the $\mathbb{PR}^*$ procedure, but a dual view of it that can further reduce the number of constraints. We describe this dual view next.

### 4.2 RESTRICTED REGION PRUNING

Just as a vector can be dominated over the whole belief space $\mathcal{B}$, a vector $u \notin \mathcal{U}$ is dominated by $\mathcal{U}$ over some belief sub region $\tilde{\mathcal{B}}$ if $\forall b \in \tilde{\mathcal{B}}, \exists u' \in \mathcal{U}$, such that $b \cdot u \leq b \cdot u'$.

If a vector is dominated over the full belief space, then it must also be dominated over every belief sub region. However the reverse is not true. We can test if a vector $w \notin \mathcal{W}$ is dominated by $\mathcal{W}$ over a belief sub region $\mathcal{B}_\mathcal{U}^u$, by solving the linear program in the procedure LP-DOMINATE[2] as listed in Table 3. We call this LP a *region-based* LP. (Note the similarity between the region-based LP and the intersection LP). Using this LP, a *region-based* pruning operator, $\mathbb{PR}(\mathcal{B}_\mathcal{U}^u, \mathcal{W})$, removes vectors that are dominated over the region $\mathcal{B}_\mathcal{U}^u$ from the set $\mathcal{W}$, in a similar way to White and Lark's technique (White 1991). The benefit of the region-based LP comes from the fact that:

$$\mathbb{PR}(\mathcal{U} \oplus \mathcal{W}) = \bigcup_{u \in \mathcal{U}} \{\{u\} \oplus \mathbb{PR}(\mathcal{B}_\mathcal{U}^u, \mathcal{W})\}. \quad (9)$$

We call this computation the *region-based* pruning of $\mathcal{U} \oplus \mathcal{W}$. Compared to the previous intersection-based pruning, which tests arbitrary pairs of regions $\mathcal{B}_\mathcal{U}^u$ and $\mathcal{B}_\mathcal{W}^w$, the region-based pruning fixes one region $\mathcal{B}_\mathcal{U}^u$, and find all regions defined by $\mathcal{W}$ that intersect with it. The benefit is that the pruning now builds LPs that have constraints ranging from $|\mathcal{U}|$ to $|\mathcal{U}| + |\mathcal{W}|$, instead of fixed at $|\mathcal{U}| + |\mathcal{W}|$.

The *restricted region* (RR) algorithm captures exactly this property (and hence the name), but it was implemented in a smart way that does not require actually setting up a different LP. Instead, by careful bookkeeping of the source from which a vector $(u+w) \in \mathcal{U} \oplus \mathcal{W}$ is constructed from, the RR algorithm is able to perform the pruning with an LP that is equivalent to a region-based LP, by simply modifying the $\mathcal{D}$ set used in step 16 of Table 1. In particular, when testing if a vector $(u+w)$ is dominated, one can choose from

$$\mathcal{D}' := (\{u\} \oplus \mathcal{W}) \cup \{u' + w | u' + w \in \mathcal{D}\}, \text{ or}$$
$$\mathcal{D}'' := (\mathcal{U} \oplus \{w\}) \cup \{u + w' | u + w' \in \mathcal{D}\},$$

to replace $\mathcal{D}$. Choosing either one resulted in the RR algorithm, while choosing the smallest among $\mathcal{D}, \mathcal{D}'$ and $\mathcal{D}''$ is called the *generalized incremental pruning* (GIP) algorithm (Cassandra, Littman, & Zhang 1997).

However, by doing so the RR algorithm also limits the savings in the number of constraints to a single cross-sum. When embedded in IP, where more than two sets are cross-summed, the savings begin to diminish when measured against the size of the input. To see this, consider pruning $\mathcal{U} \oplus \mathcal{V} \oplus \mathcal{W}$ as $\mathbb{PR}(\mathcal{U} \oplus \mathbb{PR}(\mathcal{V} \oplus \mathcal{W}))$. For $\mathcal{V} \oplus \mathcal{W}$, RR is able to achieve a worst case number of constraints of $|\mathcal{V}| + |\mathcal{W}|$ in each LP. But when pruning $\mathcal{U} \oplus \mathbb{PR}(\mathcal{V} \oplus \mathcal{W})$, the worst case number of constraints becomes $|\mathcal{U}| + |\mathbb{PR}(\mathcal{V} \oplus \mathcal{W})|$. In general, as IP progresses, the number of constraints approaches $|\mathcal{V}^a|$, which in the worst case is on the order of $\prod_z |\mathcal{V}^{a,z}|$.

---

[2]To avoid adding more symbols, we use the same names for the procedures LP-DOMINATE and $\mathbb{PR}$ as in Table 1, and let the number of parameters distinguish them.

## 5 REGION-BASED IP

In this section, we show that by explicitly constructing region-based LPs for the pruning, combined with its dual view of intersecting belief sub regions, we are able to extend their benefit across multiple cross-sums. To simplify the notation, we drop the $a$ and $z$ superscripts in Equation (3), and refer to the computation as

$$\mathcal{V} = \mathbb{PR}(\oplus_{i=1}^k \mathcal{V}_i). \quad (10)$$

Furthermore, we omit specifying the range of $i$ when the range is from 1 to $k$.

### 5.1 INTERSECTION-BASED INCREMENTAL PRUNING (IBIP)

First, we note that the intersection view captured by Theorem 1 can be extended to multiple cross-sums:

**Theorem 2** *Let $v_i \in \mathcal{V}_i$, $i = 1, \ldots, k$, then $\sum_i v_i \in \mathbb{PR}(\oplus_i \mathcal{V}_i)$ if and only if $\cap_i \mathcal{B}_{\mathcal{V}_i}^{v_i} \neq \phi$.*

Thus, the problem of computing $\mathbb{PR}(\oplus_i \mathcal{V}_i)$ is equivalent to finding all intersecting regions. The linear program listed in the procedure LP-INTERSECT (Table 2) can be easily extended to handle multiple regions, by adding additional region constraints to it. From now on we allow the LP-INTERSECT procedure to accept arbitrary number of regions as parameters. We introduce the operator $\mathbb{I}(\{\mathcal{V}_i\})$ that takes as input a set of vector sets and produces a list of intersecting regions defined by those vector sets:

$$\mathbb{I}(\mathcal{V}_{i_1}, \ldots, \mathcal{V}_{i_t}) = \left\{ (\mathcal{B}_{\mathcal{V}_{i_1}}^{v_1}, \ldots, \mathcal{B}_{\mathcal{V}_{i_t}}^{v_t}) | \cap_{j=1}^t \mathcal{B}_{\mathcal{V}_{i_j}}^{v_j} \neq \phi \right\}$$

Pruning of the cross-sums can then be expressed as

$$\mathbb{PR}(\oplus_i \mathcal{V}_i) = \left\{ \sum_i v_i \middle| (\mathcal{B}_{\mathcal{V}_1}^{v_1}, \ldots, \mathcal{B}_{\mathcal{V}_k}^{v_k}) \in \mathbb{I}(\mathcal{V}_1, \ldots, \mathcal{V}_k) \right\} \quad (11)$$

A naive approach to compute $\mathbb{I}(\{\mathcal{V}_i\})$ is to enumerate all possible combinations of $\{\mathcal{B}_{\mathcal{V}_i}\}$, and test them for intersection using the intersection LP. This requires a total of $\prod_i |\mathcal{V}_i|$ LPs, but each LP has only $\sum_i |\mathcal{V}_i|$ constraints. A better approach would be to use an incremental process similar to IP: To compute $\mathbb{I}(\mathcal{V}_1, \mathcal{V}_2, \ldots, \mathcal{V}_k)$, we test if LP-INTERSECT($\mathcal{B}_{\mathcal{V}_1}, \mathcal{B}_{\mathcal{V}_2}, \ldots, \mathcal{B}_{\mathcal{V}_k}$) is true for all combinations of $\mathcal{B}_{\mathcal{V}_1}$ and $(\mathcal{B}_{\mathcal{V}_2}, \ldots, \mathcal{B}_{\mathcal{V}_k})$, where $(\mathcal{B}_{\mathcal{V}_2}, \ldots, \mathcal{B}_{\mathcal{V}_k}) \in \mathbb{I}(\mathcal{V}_2, \ldots, \mathcal{V}_k)$, and $\mathbb{I}(\mathcal{V}_2, \ldots, \mathcal{V}_k)$ is computed recursively in the same manner. The recursion stops at $\mathbb{I}(\mathcal{V}_{k-1}, \mathcal{V}_k)$, at which point the naive approach is used to compute the results. We call this algorithm for computing $\mathbb{I}$ and subsequently $\mathbb{PR}(\oplus_i \mathcal{V}_i)$ the *intersection-based incremental pruning* (IBIP).

Surprisingly, IBIP solves the exact same number of LPs as IP (and the RR variants). To see this, consider the top level of the recursion. The total number of combinations between $\mathcal{B}_{\mathcal{V}_1}$ and $(\mathcal{B}_{\mathcal{V}_2}, \ldots, \mathcal{B}_{\mathcal{V}_k})$, and hence the number of LPs needed, is

$$|\mathcal{V}_1| \times |\mathbb{I}(\mathcal{V}_2, \ldots, \mathcal{V}_k)| = |\mathcal{V}_1| \times |\mathbb{PR}(\oplus_{i=2}^k \mathcal{V}_i)|,$$

which is also the number of LPs needed at the top recursion of IP (see end of Section 3). Similarly the same numbers of LPs are solved at all recursive steps. It follows that the total numbers of LPs of the two approaches are the same: $|\mathcal{V}| \sum |\mathcal{V}_i|$.

However, all the LPs used in computing $\mathbb{I}$ have at most $\sum_{i=1}^k |\mathcal{V}_i|$ constraints. In particular, when computing $\mathbb{I}(\mathcal{V}_t, \ldots, \mathcal{V}_k)$, the number of constraints ranges from $\sum_{i=t+1}^k |\mathcal{V}_i|$ to $\sum_{i=t}^k |\mathcal{V}_i|$. Thus, to compute $\mathbb{PR}(\oplus_i \mathcal{V}_i)$, the IBIP algorithm requires the same number of LPs but with possibly an exponential reduction in the number of constraints compared to IP. The number of constraints does not depend on the size of the output set, as with IP.

## 5.2 REGION-BASED INCREMENTAL PRUNING (RBIP)

In this section, we show how the number of constraints in IBIP can be further reduced, in a way similar to how RR reduces the number of constraints from the intersection-based pruning in the single cross-sum case (Section 4.2). To make a direct comparison with the recursion in IBIP, we will start from $\mathcal{V}_k$: To compute $\mathbb{I}(\mathcal{V}_1, \mathcal{V}_2, \ldots, \mathcal{V}_k)$, we first fix a region in $\mathcal{V}_k$, call it $\mathcal{B}_{\mathcal{V}_k}$, and find all the elements in $\mathbb{I}(\mathcal{V}_1, \ldots, \mathcal{V}_{k-1})$ that intersect with $\mathcal{B}_{\mathcal{V}_k}$. We repeat this for all the regions in $\mathcal{V}_k$.

To find all the regions in $\mathbb{I}(\mathcal{V}_1, \ldots, \mathcal{V}_{k-1})$ that intersect with $\mathcal{B}_{\mathcal{V}_k}$, we first find all regions in each $\mathcal{V}_i (1 \leq i \leq k-1)$ that intersect with $\mathcal{B}_{\mathcal{V}_k}$. Recall that each such region corresponds to a vector in the vector set, and the set of intersecting regions corresponds to some subset of vectors $\mathcal{V}'_i \subseteq \mathcal{V}_i$. $\mathcal{V}'_i$ can be precisely computed by the region-based pruning, $\mathcal{V}'_i = \mathbb{PR}(\mathcal{B}_{\mathcal{V}_k}, \mathcal{V}_i)$. Once all $\mathcal{V}'_i$ are computed, we then recursively compute $\mathbb{I}(\mathcal{V}'_1, \ldots, \mathcal{V}'_{k-1})$, by fixing a $\mathcal{B}_{\mathcal{V}'_{k-1}}$ and find all the elements in $\mathbb{I}(\mathcal{V}'_1, \ldots, \mathcal{V}'_{k-2})$ that intersect $\mathcal{B}_{\mathcal{V}_k} \cap \mathcal{B}_{\mathcal{V}'_{k-1}}$. Note that the $\mathbb{I}$ operator serves only as a conceptual place-holder in this process; all computations are carried out using the region-based pruning operator.

Table 4 shows the algorithm that finds the set of intersecting regions using this process. We call the algorithm that computes $\mathbb{PR}(\oplus_i \mathcal{V}_i)$ using Table 4 and Equation (11) the *region-based incremental pruning* (RBIP) algorithm.

The main motivation for RBIP is to further reduce the number of constraints. As Table 4 shows, all pruning in RBIP is of the form $\mathbb{PR}(\tilde{\mathcal{B}}, \mathcal{V}_t)$. In line 3, the pruning corresponds to testing some $\mathcal{B}_{\mathcal{V}_1}$ with some $(\mathcal{B}_{\mathcal{V}_2}, \ldots, \mathcal{B}_{\mathcal{V}_k})$ for intersection in IBIP. The number of constraints in IBIP is from $\sum_{i=2}^k |\mathcal{V}_i|$ to $\sum_{i=1}^k |\mathcal{V}_i|$. The number of constraints in

Table 4: Region-based pruning for computing $\mathbb{I}$.

---

**prpcedure** $\mathbb{I}^*(\tilde{\mathcal{B}}, \{\mathcal{V}_i | i \in [1, t]\})$
1.   $\mathcal{K} \leftarrow \phi$
2.   if $t = 1$
3.     $\mathcal{K} \leftarrow \{\mathcal{B}_{\mathcal{V}_1}^v | v \in \mathbb{PR}(\tilde{\mathcal{B}}, \mathcal{V}_1)\}$
4.   else
5.     for each $v \in \mathcal{V}_t$
6.       $\mathcal{V}'_i \leftarrow \mathbb{PR}(\tilde{\mathcal{B}} \cap \mathcal{B}_{\mathcal{V}_t}^v, \mathcal{V}_i), i \in [1, t-1]$
7.       if $\exists i \in [1, t-1]$ such that $\mathcal{V}'_i = \phi$
8.         continue
9.       $\mathcal{D} \leftarrow \mathbb{I}^*(\tilde{\mathcal{B}} \cap \mathcal{B}_{\mathcal{V}_t}^v, \{\mathcal{V}'_i | i \in [1, t-1]\})$
10.      $\mathcal{K} \leftarrow \mathcal{K} \cup \{(\mathcal{B}_{\mathcal{V}_1}, \ldots, \mathcal{B}_{\mathcal{V}_{t-1}}, \mathcal{B}_{\mathcal{V}_t}^v) | (\mathcal{B}_{\mathcal{V}_1}, \ldots, \mathcal{B}_{\mathcal{V}_{t-1}}) \in \mathcal{D}\}$
11.  return $\mathcal{K}$
**procedure** $\mathbb{I}(\mathcal{V}_1, \mathcal{V}_2, \ldots, \mathcal{V}_k)$
12.  return $\mathbb{I}(\mathcal{B}, \{\mathcal{V}_i | i \in [1, k]\})$

---

RBIP ranges between $\sum_{i=2}^k |\mathcal{V}_i^*|$ and $\sum_{i=1}^k |\mathcal{V}_i^*|$, where $\mathcal{V}_i^*$ is $\mathcal{V}_i$ pruned multiple times previously at line 6. Because of the region-based pruning, $|\mathcal{V}_i^*|$ could be much smaller than $|\mathcal{V}_i|$ and this is where the savings come from. The analysis of the pruning at line 6 follows similarly.

In theory, it is possible that the region-based pruning at line 6 may not prune any vector at all. In this case there is no saving in the number of constraints as compared to IBIP. Further, in this worst case scenario, the number of LPs solved by RBIP is $|Z||\mathcal{V}| \sum |\mathcal{V}_i|$, or $|Z|$ times that of IBIP. It remains an open question whether this happens in realistic POMDPs. In all the experiments we have performed so far, we observed substantial savings in terms of both the number of LPs and the number of constraints using RBIP. We present these results next.

## 6 EXPERIMENTAL RESULTS

In this section, we present experimental results comparing the performance of IBIP, RBIP, and the GIP algorithms (Section 4). GIP performs uniformally better than RR in all our tests. We used the POMDP code by Cassandra (1999) as the basis of our implementation, and used his implementation of the GIP algorithm for comparison.

### 6.1 PROBLEMS FROM THE LITERATURES

We first tested the algorithms on a set of problems from the literature that are publicly available from (Cassandra 1999). These problems are listed in Table 5. Note that our algorithm addresses the exponential blow-up associated with the number of observations. With 2 observations, our algorithms are essentially the same as the GIP algorithm. Thus we chose problems that have more than 2 observations. In many such problems, the number of observations in the problem description is usually larger than the number of *actual* observations – observations that are possible in any given state. This *actual* number is the number of vec-tor sets whose cross-sum needs to be pruned. We show the number of observations in the problem in column "$Z$", and the maximal number of actual observations for all actions in column "$Z^*$". Column "T" is the number of iterations of DP ran to collect the data. Only data for the pruning of the cross-sums are shown, because that is the only part affected by our algorithms. The column under "time" is the time spent on the pruning of the cross-sums, in CPU seconds. The column under "LP" is the total number of linear programs solved during the pruning, in $10^3$, and the column under "C" is the total number of constraints in those linear programs, in $10^6$. A limit of 10 hours was set for all the algorithms in these tests, after which they were terminated.

Table 5: Test results on problems from the literature. Times are shown in seconds except when otherwise noted.

| problem | $Z$ | $Z^*$ | T | time | LP | C |
|---|---|---|---|---|---|---|
| 4x3 | 6 | 2 | 10 | 19.50 | 14.38 | 0.78 |
|  |  |  |  | 11.91 | 14.38 | 0.78 |
|  |  |  |  | 11.11 | 14.38 | 0.78 |
| shuttle | 5 | 3 | 10 | 9.86 | 10.07 | 0.59 |
|  |  |  |  | 7.58 | 10.07 | 0.56 |
|  |  |  |  | 7.65 | 10.41 | 0.56 |
| Maze20 | 8 | 4 | 3 | >10hr | na | na |
|  |  |  |  | 30649.68 | 3545.58 | 1559.97 |
|  |  |  |  | 6540.66 | 914.09 | 238.43 |
| iff | 22 | 20 | 2 | >10hr | na | na |
|  |  |  |  | 400.11 | 608.41 | 22.00 |
|  |  |  |  | 329.07 | 759.32 | 12.45 |

Table 6: Results on problem sets $(k, 10)$. For $k > 6$ only data for IBIP and RBIP is available.

| $k$ | $|\mathcal{V}|$ | time | LP | C |
|---|---|---|---|---|
| 2 | 67 | 0.02 | 0.08 | 0.0011 |
|  | 67 | 0.01 | 0.10 | 0.0014 |
|  | 67 | 0.02 | 0.10 | 0.0014 |
| 3 | 365 | 0.37 | 0.71 | 0.02 |
|  | 367 | 0.17 | 0.77 | 0.01 |
|  | 367 | 0.17 | 0.71 | 0.01 |
| 4 | 1740 | 10.63 | 4.44 | 0.40 |
|  | 1740 | 1.37 | 4.44 | 0.11 |
|  | 1740 | 1.22 | 3.77 | 0.09 |
| 5 | 5788 | 209.32 | 21.84 | 5.87 |
|  | 5788 | 7.20 | 21.84 | 0.65 |
|  | 5788 | 5.41 | 14.33 | 0.41 |
| 6 | 21884 | 3665.81 | 79.72 | 71.58 |
|  | 21883 | 31.72 | 79.72 | 2.74 |
|  | 21653 | 25.66 | 54.90 | 1.87 |
| 7 | 60845 | 127.42 | 298.55 | 11.83 |
|  | 59886 | 92.39 | 171.32 | 6.61 |
| 8 | 119789 | 393.62 | 907.00 | 39.45 |
|  | 117596 | 231.72 | 399.62 | 16.74 |
| 9 | 318583 | 1043.15 | 2104.89 | 97.29 |
|  | 310737 | 716.22 | 1117.77 | 51.22 |
| 10 | 720501 | 2738.04 | 5290.72 | 263.87 |
|  | 703278 | 1802.65 | 2590.10 | 126.78 |

For each problem, we list the results for the GIP algorithm in the top row, followed by IBIP and then RBIP. As we can see, for the 4x3 problem, where only 2 sets of vectors are cross-summed, there is no difference in the number of LPs and the number of constraints. Even so, the time used by GIP is slightly longer. We conjecture that this is due to the different LP formations used by the different algorithms.

For the problem "shuttle", where there are 3 actual observations, our algorithms begin to show their advantage in terms of the number of constraints. For the two larger problems "Maze20" and "iff", GIP cannot finish the cross-sum pruning for all actions within the 10 hour limit. On "Maze20", GIP did not finish the pruning of a single set of the cross-sums involving 4 vector sets. On "iff", GIP did finish the pruning of 2 of the 4 sets of cross-sums at the end of the 10-hour period (36572.52 seconds). Thus RBIP is at least 110 times faster than GIP on this problem. For the 2 sets processed, GIP solved $61.14 \times 10^3$ LPs, with a total constraints of $28.96 \times 10^6$, which is already more than twice the total number of constraints by RBIP on all 4 sets.

On the two larger problems, RBIP uses significantly fewer constraints than IBIP. This is due to the region-based pruning. In "shuttle" and "iff", RBIP needs to solve slightly more LPs than IBIP, while in "Maze20", RBIP solves significantly fewer LPs than IBIP. In all cases, RBIP is at least as fast as IBIP, and in many cases a lot faster.

## 6.2 ARTIFICIAL PROBLEMS

Because of the lack of data collected for GIP on the larger problems from the literature, we present a special experiment to better demonstrate the scalability of our algorithms. We construct a set of random vector sets $\{\mathcal{V}_1, \ldots, \mathcal{V}_k\}$ and feed it directly to the algorithms to compute $\mathbb{PR}(\oplus_{i=1}^{k} \mathcal{V}_i)$. This way, we can easily vary the number of sets $k$ and the size of the input sets involved in the cross-sum.

Random vector sets, all with 10 states, are created as follows. The set $\mathcal{V}_i$ is initialized with a random vector, which is generated by drawing 10 numbers uniformaly from $[-100.0, 100.0]$. Then, additional random vectors are generated and added to the set provided that they are not dominated. The procedure LPDOMINATE$(v, \mathcal{V}_i)$ is used to determine if a new vector $v$ is dominated. This process is repeated until the number of vectors in $\mathcal{V}_i$ reaches $n$. A test problem is thus specified by the pair $(k, n)$.

It is hard to to determine whether vector sets created this way represent vector sets encountered in typical POMDPs. However we do note that it is easy to "reverse-engineer" a POMDP given an arbitrary set of vectors such that after one step of DP the exact same vectors are created.

Table 6 lists detailed results comparing the three algorithms with respect to a problem set $(k, 10)$, where $k$ ranges from 2 to 10. Column "$|\mathcal{V}|$" is the size of the resulting set af-ter the pruning. As we can see clearly from the table, our two algorithms greatly outperform the GIP algorithm, especially for larger $k$ values. For $k = 6$, IBIP is about 120 faster, and RBIP more than 140 times faster than GIP. The GIP algorithm can not finish within 10 hours for any of the problems with $k > 6$.

Table 7: Speed-up factors compared to GIP on a range of problems $(k, n)$.

|   | $n$ | | | | |
|---|---|---|---|---|---|
| $k$ | 15 | 20 | 25 | 30 | 35 |
| 2 | 1.17 | 1.21 | 1.10 | 1.23 | 1.14 |
|   | 1.17 | 1.21 | 1.14 | 1.23 | 1.13 |
| 3 | 2.98 | 4.08 | 4.53 | 4.78 | 4.22 |
|   | 3.28 | 4.83 | 5.96 | 7.02 | 8.46 |
| 4 | 18.23 | 18.99 | 23.48 | 17.72 | 11.50 |
|   | 23.35 | 30.78 | 31.98 | 53.85 | 39.27 |
| 5 | 78.28 | 81.77 | [216.39] | [476.60] | [680.51] |
|   | 131.49 | 123.27 | [83.21] | [111.63] | [171.11] |
| 6 | [200.53] | [456.26] | [669.45] | [2009.74] | [3696.99] |
|   | [143.10] | [176.36] | [238.38] | [433.55] | [655.01] |

We note that for some test problems, the different algorithms give slightly different results as indicated by the $|\mathcal{V}|$ column. This is due to accumulated numerical errors in solving the LPs, and is more noticeable for larger problems.

Finally, Table 7 presents data showing the speed-up factor compared to GIP on a range of problems $(k, n)$. The numbers in the table show the ratio between the time used by GIP for the pruning, and the time used by IBIP and RBIP. For each problem, data for IBIP is shown on the top row, and data for RBIP is shown on the bottom row. Numbers in brackets are actual running times (in CPU seconds) for the algorithms; no data is available from GIP to compute the speed-up factor on these problems, because GIP did not finish after 10 hours on those problems. From this table, we can see that the performance gain is much more dramatic along $k$, the number of vector sets, than along $n$, the size of each set.

## 7 CONCLUSIONS

We have presented two new algorithms, IBIP and RBIP, for the pruning of cross-sums in the dynamic programming update for POMDPs. Our algorithms build upon the original IP algorithm and its RR variants. While IP and RR need to solve an exponential number of LPs with exponentially many constraints, our algorithms solve roughly the same amount of LPs, but with only a polynomial number of constraints. Our algorithms demonstrated over 100 times performance speed-up on some problems from the literature, as well as a set of artificial test problems.

POMDPs are extremely difficult to solve, and the cross-sum is only one of the several exponential components in the whole DP process. We are working on other exponential components and hope to further improve the scalability of the DP step.

**Acknowledgment** This work was supported in part by the National Science Foundation under grant IIS-0219606, and by the Air Force Office of Scientific Research under grant F49620-03-1-0090. Any opinions, findings, and conclusions or recommendations expressed in this material are those of the authors and do not reflect the views of the NSF or AFOSR.

## References

Cassandra, A.; Littman, M.; and Zhang, N. 1997. Incremental pruning: A simple, fast, exact method for partially observable markov decision processes. In *Proceedings of the 13th Annual Conf. on Uncertainty in Artificial Intelligence (UAI-97)*, 54–61.

Cassandra, A. R. 1998. *Exact and Approximate Algorithms for Partially Observable Markov Decision Processes*. Ph.D. Dissertation, Brown University.

Cassandra, A. R. 1999. Tony's POMDP page. http://www.cs.brown.edu/research/ai/pomdp/.

Feng, Z., and Hansen, E. 2001. Approximate planning for factored POMDPs. In *Proceedings of the 6th European Conference on Planning*.

Hansen, E. A., and Feng, Z. 2000. Dynamic programming for POMDPs using a factored state representation. In *Proceedings of the 5th international conference on Artificial Intelligence Planning & Scheduling*.

Hansen, E. A. 1998. An improved policy iteration algorithm for partially observable MDPs. In *Proceedings of the 14th conference on uncertainty in Artificial Intelligence (UAI-98)*.

Littman, M.; Cassandra, A.; and Kaelbling, L. 1996. Efficient dynamic-programming updates in partially observable markov decision processes. Technical Report CS-95-19, Brown University, Providence, RI.

Littman, M. 1994. The witness algorithm: Solving partially observable markov decision processes. Technical Report CS-94-40, Brown University Department of Computer Science.

Smallwood, R., and Sondik, E. 1973. The optimal control of partially observable Markov processes over a finite horizon. *Operations Research* 21:1071–1088.

White, C. 1991. A survey of solution techniques for the partially observed markov decision process. *Annals of Operations Research* 32:215–230.

Zhang, N. L., and Liu, W. 1996. Planning in stochastic domains: Problem characteristics and approximation. Technical Report HKUST-CS96-31, Hong Kong University of Science and Technology.